\documentclass[times,twocolumn,final]{article}

\usepackage{authblk}
\usepackage[backend=bibtex,style=numeric-verb]{biblatex}
\usepackage[utf8]{inputenc}
\usepackage[english]{babel}

\usepackage{framed,multirow}

\usepackage{amssymb}
\usepackage{latexsym}

\usepackage{url}
\usepackage{xcolor}
\definecolor{newcolor}{rgb}{.8,.349,.1}

\usepackage{graphicx,subcaption}
\usepackage{lineno,hyperref}
\usepackage{tabularx}
\usepackage{booktabs}
\usepackage{siunitx}
\usepackage{textcomp}
\modulolinenumbers[5]

\usepackage{url}

\usepackage{breakurl}

\bibliography{references}

\begin{document}

\thispagestyle{empty}

\title{Semantic Segmentation on Swiss3DCities: A Benchmark Study on Aerial Photogrammetric 3D Pointcloud Dataset}

\author[1]{G\"{u}lcan Can \thanks{Corresponding Author: gulcan@nomoko.world}}

\author[2]{Dario Mantegazza \thanks{dario.mantegazza@idsia.ch}}

\author[2]{Gabriele Abbate \thanks{gabriele.abbate@idsia.ch}}

\author[1]{S\'{e}bastien Chappuis \thanks{sebastien@nomoko.world}}

\author[2]{Alessandro Giusti \thanks{alessandrog@idsia.ch}}

\affil[1]{Nomoko AG, Zurich, Switzerland}
\affil[2]{The Dalle Molle Institute for Artificial Intelligence (IDSIA), Viganello, Switzerland}

\date{\vspace{-5ex}}
\maketitle

\begin{abstract}
We introduce a new outdoor urban 3D pointcloud dataset, covering a total area of \SI{2.7}{\km \squared}, sampled from three Swiss cities with different characteristics. The dataset is manually annotated for semantic segmentation with per-point labels, and is built using photogrammetry from images acquired by multirotors equipped with high-resolution cameras.  In contrast to datasets acquired with ground LiDAR sensors, the resulting point clouds are uniformly dense and complete, and are useful to disparate applications, including autonomous driving, gaming and smart city planning.  As a benchmark, we report quantitative results of PointNet++, an established point-based deep 3D semantic segmentation model; on this model, we additionally study the impact of using different cities for model generalization.
\end{abstract}

\section{Introduction}
\label{sec:intro}

\begin{table*}[tp!]
	\small
	\centering
	\caption{Comparison of our datasets with the common 3D pointcloud datasets in the literature. The spatial size denotes either the covered area or the driven length (in the case of mobile LiDAR captures) for a dataset.}
	\label{tbl:relatedWork}
	\begin{tabular}{ccrcccc}
		\toprule
		\textbf{Paper}                       &\multicolumn{1}{c}{\textbf{\begin{tabular}[c]{@{}c@{}}Data\\Acquisition\end{tabular}}} 
		
		&\multicolumn{1}{c}{\textbf{\begin{tabular}[c]{@{}c@{}}Spatial\\Size\end{tabular}}}                                                                                                                 
		&\textbf{City}
		&\textbf{Cat.}                                                   
		
		&\textbf{RGB}
		
		& \multicolumn{1}{c}{\textbf{Points}} 
		\\ \midrule
		
		\multicolumn{1}{c}{\begin{tabular}[c]{@{}c@{}}Semantic\\KITTI\cite{behley2019semantickitti}\end{tabular}}                & Mobile LiDAR                             & $39.2\times10^3 m$ &  Karlsruhe  &  25 & No & 4549M                                                                                  
		\\ \cline{1-7} 
		Paris-Lille3D\cite{roynard2018parislille3d}                 & Mobile LiDAR                             & $1.94\times10^3 m$ & \multicolumn{1}{c}{\begin{tabular}[c]{@{}c@{}}Paris\\Lille\end{tabular}} & 9 & No & 143M                                                                                    \\ \cline{1-7} 
		Toronto3D\cite{tan2020toronto3d}                 & Mobile LiDAR                             & $1\times10^3 m$ & Toronto    & 8 & Yes & 78.3M                                                                                       
		\\ \cline{1-7} 
		Semantic3D\cite{semantic3d_hackel2017isprs}                 & Static LiDAR                             & -\quad\quad\quad & St. Gallen    & 8 & Yes & 4009M                                                                                          
		\\ \cline{1-7} 
		ISPRS\cite{niemeyer2014contextual}                 & Aerial LiDAR                             & -\quad\quad\quad & Vaihingen   & 9 & No & 1.2M                                                                                            
		\\ \cline{1-7} 
		DublinCity\cite{zolanvari2019dublincity}                 & Aerial LiDAR                             & $2\times10^6$\SI{}{\m\squared} & Dublin    & 13 & No & 260M   
		\\ \cline{1-7}
		
        DALES\cite{varney2020dales}                 & Aerial LiDAR                             & $10\times10^6$\SI{}{\m\squared} & Surrey, BC   &  8 & No & 505M     
        \\ \cline{1-7} 
        \multicolumn{1}{c}{\begin{tabular}[c]{@{}c@{}}SenSat\\Urban\cite{hu2020sensaturban}\end{tabular}}                 & \multicolumn{1}{c}{\begin{tabular}[c]{@{}c@{}}UAV\\Photogrammetry\end{tabular}}                             & $6 \times 10^6$\SI{}{\m\squared} & \multicolumn{1}{c}{\begin{tabular}[c]{@{}c@{}}Birmingham\\Cambridge\end{tabular}}    & 13 & Yes & 2847M                                                                                       
        \\ \cline{1-7} 
		\multicolumn{1}{c}{\textbf{\begin{tabular}[c]{@{}c@{}}Swiss3DCities\\(this paper)\end{tabular}}}                  &   \multicolumn{1}{c}{\begin{tabular}[c]{@{}c@{}}UAV\\Photogrammetry\end{tabular}}                          & $2.7 \times 10^6$\SI{}{\m\squared} & \multicolumn{1}{c}{\begin{tabular}[c]{@{}c@{}c@{}}Zurich\\Zug\\Davos\end{tabular}} &  5 & Yes & \multicolumn{1}{c}{\begin{tabular}[c]{@{}c@{}c@{}}226M\\(Sparse: 7.5M \\Dense: 3147M)\end{tabular}}                                                                         
		\\ \bottomrule
	\end{tabular}
\end{table*}

Many recent achievements of deep learning depend on the availability of very large labeled training datasets~\cite{schmidhuber2015deep, goodfellow2016deep}, such as ImageNet~\cite{imagenet} for image classification and MS COCO~\cite{MSCOCO} for image segmentation.  In this work, we propose a new dataset of dense urban 3D pointclouds, spanning \SI{2.7}{\km\squared}, acquired using photogrammetry from three different cities in Switzerland (Zurich, Zug and Davos).  The entire dataset is manually annotated with dense labels, which associate each point to one of five categories: terrain, construction, vegetation, vehicle, and urban asset.

The main goal of the dataset is to train and validate semantic segmentation algorithms for urban environments. Semantic segmentation consists in partitioning the data into multiple sets of points, such that each set represents only objects of a given type. The problem is relevant for many real-world applications, such as autonomous or assisted driving, automated content generation for games~\cite{hendrikx2013procedural}, augmented and virtual reality applications, and city planning~\cite{yang2019developing}. 

Most existing datasets~\cite{roynard2018parislille3d,behley2019semantickitti,tan2020toronto3d} for outdoor 3D semantic segmentation are motivated by real-time autonomous driving applications, and are therefore acquired at low resolution by street-level Light Detection and Ranging (LiDAR) sensors; this yields incomplete point clouds (for example, areas far from roads, such as roofs, are either not acquired or acquired with very low resolution) which are unsuitable for applications such as city planning, urban augmented or virtual reality (AR/VR), or gaming.  In contrast, we acquire high-resolution photographs from unmanned aerial vehicles (UAV) flying a grid pattern over the area of interest, then reconstruct the 3D shape using photogrammetry; this allows us to densely acquire most outdoor surfaces.  Similar approaches have been previously adopted for several applications, including automatic urban area mapping~\cite{nex2014uav3Dmapping}, damage detection~\cite{mohammadi2019damage}, and cultural heritage site mapping for digital preservation~\cite{poux20173darchaeology}. Compared to 3D models built by satellite-borne cameras, this approach yields models with higher-resolution geometry and texture.

High resolution data yields more accurate models, but also aids the segmentation task because it contains more information to discriminate between different classes;
currently, state-of-the-art models for 3D semantic segmentation rely on deep learning~\cite{guo2019deeplearning3Dpointclouds, zhang2019deeplearning3Dpointclouds, bello2020deeplearning3Dpointclouds} and represent input data as voxels~\cite{zhou2018voxelnet}, points~\cite{qi2017pointnet} or meshes~\cite{le2017multi}; other approaches render multiple views of the 3D scene and then rely on 2D semantic segmentation models~\cite{su2015MVCNN, su2018splatnet, boulch2018snapnet}, which can be trained on more abundant 2D labeled semantic segmentation datasets.

To show the potential of our dataset for training and evaluating segmentation algorithms, we consider the well-established PointNet++ model~\cite{qi2017pointnet, qi2017pointnet++} and report its performance when using different splits for training and evaluation. In particular, the performance of machine learning models depends not only on the size of the training dataset, but also on how representative it is of the evaluation data: often, models trained on large amounts of data from a given environment fail to generalize to a different target environment.  Because our dataset contains data from three cities with different characteristics, it can be used to explore this fundamental aspect.



The rest of the paper is organized into five sections.  We first describe related commonly-used datasets for 3D semantic segmentation in Section~\ref{sec:relatedwork}. Then, in Section \ref{sec:data} we present our \textbf{main contribution}: a new pointwise labeled multi-city dataset for semantic segmentation of outdoor 3D point clouds, which we release to the research community in three versions with different point densities; we characterize the dataset and describe data acquisition, processing and manual labeling pipelines.  In Section~\ref{sec:method} we describe the applied deep learning model to demonstrate semantic segmentation task on our dataset. We discuss quantitative results in Section~\ref{sec:result}, where we also explore the model's generalization ability across different cities (\textbf{secondary contribution}). Section \ref{sec:conclusion} concludes the paper.

\begin{figure*}[tp!]
  \centering
    \includegraphics[width=0.9\textwidth]{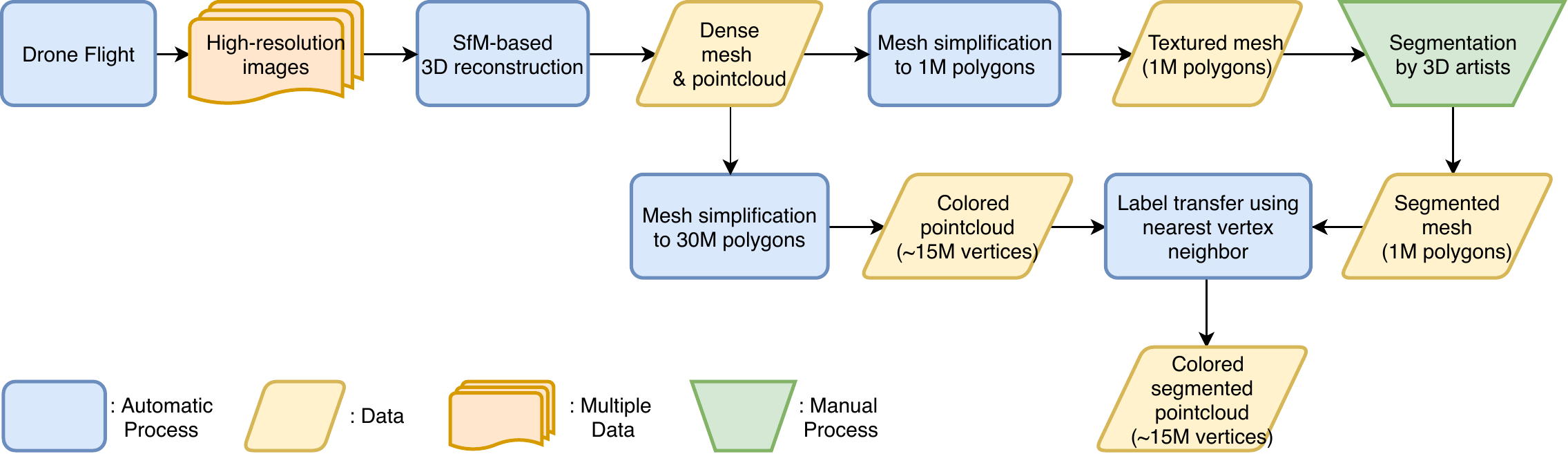}
  \caption{Our data processing and segmentation pipeline.}
  \label{fig:data_pipeline}
\end{figure*}
\section{Related Work}
\label{sec:relatedwork}
This section summarizes relevant pointcloud datasets with semantic segmentation labels (see Table \ref{tbl:relatedWork}).  One fundamental difference among the datasets is their acquisition modality, i.e. LiDAR or photogrammetry.

\subsection{LiDAR Datasets}

A lot of recent research efforts are related to autonomous driving applications: in particular, recognizing and segmenting roads and relevant urban elements from images or 3D point clouds \emph{acquired by the car itself}.  In this context, laser scanning systems, e.g. Velodyne HDL-64E~\cite{velodynlidar_hdl_64e}, are commonly used to acquire high-accuracy LiDAR pointcloud sequences from a car's point of view. Paris-Lille\cite{roynard2018parislille3d}, Semantic KITTI\cite{behley2019semantickitti}, and Toronto3D dataset\cite{tan2020toronto3d} are among such large-scale datasets with pointwise semantic labels. 

Due to the low-lying viewpoint and focus on driving-related segmentation tasks, these mobile LiDAR datasets show incomplete point clouds: e.g. the upper floors or roofs of the buildings are usually not captured. Even though these datasets serve their main scope very well, they are not suitable for other applications, such as urban planning.


Semantic3D dataset\cite{semantic3d_hackel2017isprs} is a large-scale pointcloud dataset with per-point semantic labels. This dataset is acquired via a static terrestrial laser scanning system in the north-east of Switzerland. Several points to note about this dataset are gaps due the occlusions (also known as LiDAR shadows), moving object artifacts, and varying point density based on the distance of the laser system to each surface or object in the scene.

As captured from air (either from a UAV or a helicopter), aerial LiDAR datasets such as ISPRS airborne LiDAR pointcloud dataset \cite{niemeyer2014contextual}, DublinCity dataset \cite{zolanvari2019dublincity} and DALES dataset \cite{varney2020dales} are also relevant in our context. One important difference of these datasets with respect to ours is that, due to the narrow divergence of laser beams, they can sometimes capture ground samples even when covered by vegetation. 
Compared to the DublinCity aerial LiDAR dataset\cite{zolanvari2019dublincity}, ours covers a moderately larger area and, in the medium-density version, has a similar point density. This shows the relevance of our contribution with respect to existing aerial LiDAR datasets.

\paragraph{Photogrammetric pointclouds}

Sun3D \cite{xiao2013sun3d} and Stanford Large-Scale Indoor Spaces 3D (S3DIS) \cite{armeni20163dS3DIS} are commonly used pointcloud datasets acquired using Structure-from-motion 3D reconstruction techniques. These datasets are focused on indoor scenes, and present interesting challenges for computer vision research, such as the presence of clutter, and relevant context around different objects, that can play a role in scene understanding.  Due to their limited extent, the capturing process is much less challenging than in large-scale outdoor contexts, which also need to account for variability of weather, illumination conditions, and scales of represented objects. 


The Pix4D dataset \cite{becker2018classification} comprises of aerial photogrammetric pointclouds from three outdoor scenes 
with different distributions of urban surfaces or objects. The authors emphasize the importance of color features apart from geometric features to classify these pointclouds into 6 semantic classes. This dataset is relative small-scale, since it comprises of only three scenes with a total of 18.2 million points. 

The SenSatUrban dataset \cite{hu2020sensaturban} is also reconstructed via photogrammetry from aerial photographs. The photographs are taken with a UAV that follows a double-grid flight path and covers a \SI{6}{\km\squared} area in three cities in UK (Birmingham, Cambridge, and York). Pointwise semantic labels in 13 categories are available for these pointclouds.
As an urban-focused aerial photogrammetric pointcloud, the SenSatUrban dataset is the most relevant with respect to our contribution.
SenSatUrban covers an approximately twice larger area than ours, and uses 13 categories instead of five; its point density is higher than our medium-density version, but lower than our high-density version.

\section{Dataset Description}
\label{sec:data}
In this section, we describe the process used to produce our large scale aerial photogrammetry dataset, covering both acquisition of source photographs and processing to obtain 3D point clouds. We conclude the section by detailing the data characteristics.

\subsection{Data Acquisition}

\begin{figure}[bp!]
    \centering
    \includegraphics[width=\linewidth]{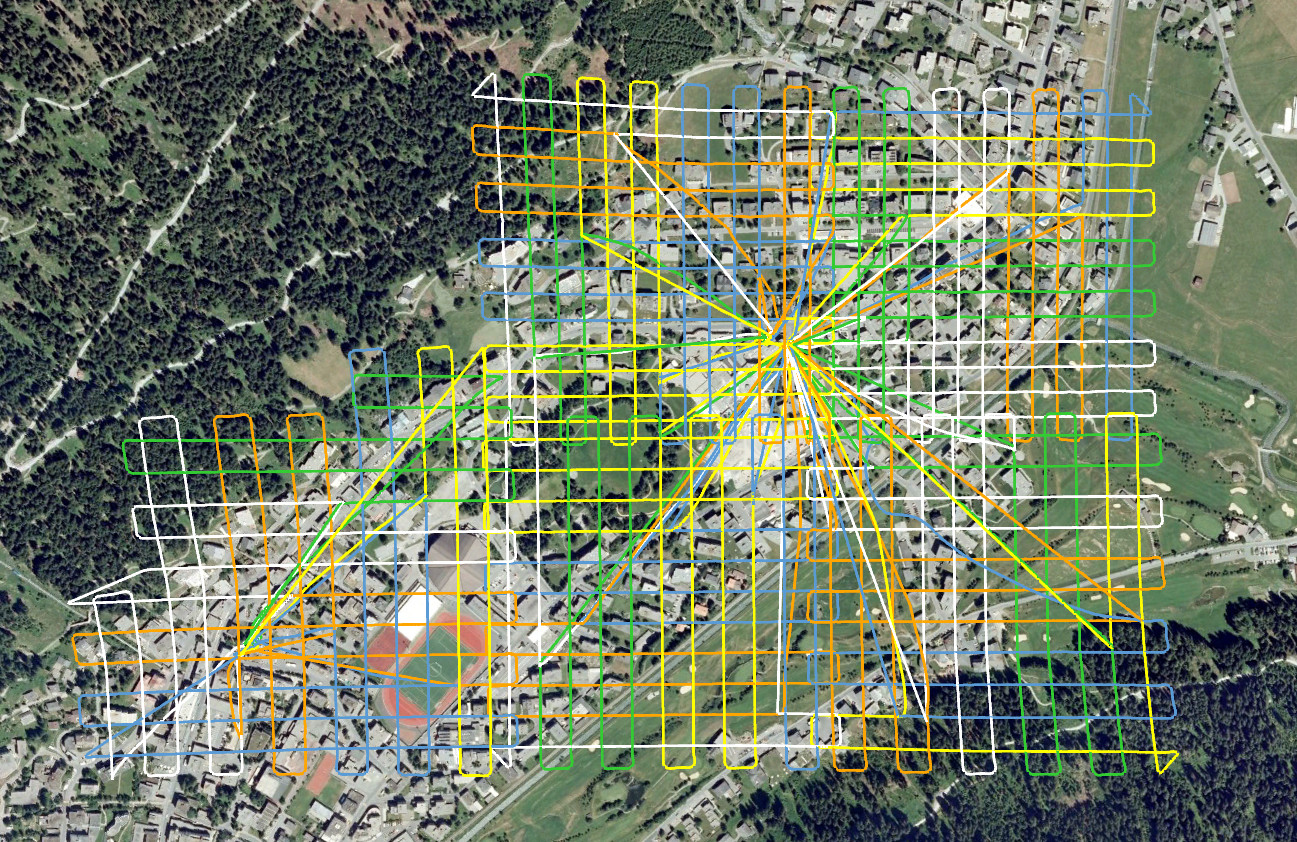}
	\caption{The flight paths while capturing the five Davos tiles.}
	\label{fig:Davos_flight_path}
	\vspace*{-3mm}
\end{figure}

The image data is acquired via a high-resolution camera array (nadir and oblique cameras) 
mounted on a 
multirotor drone. 

To capture the image data, the drone is configured to trigger the cameras simultaneously at regular intervals. 
As shown in Figure~\ref{fig:Davos_flight_path}, the drone follows a double grid flight path~\cite{rubi2019flightpathsurvey}. 


Each flight acquires one or more tiles. A single tile corresponds to \SI{412 x 412}{\m} horizontal area approximately (around 17 hectares). 
The Ground Sampling Distance (GSD), i.e. the inter-pixel distance measured on the ground, is planned as \SI{1.25}{\cm} and ultimately measured as \SI{1.28}{\cm} cm.

\subsection{Data Processing}

After aerial image acquisition, we follow a classic photogrammetry workflow to reconstruct textured 3D models, based on RealityCapture~\cite{realitycapture,CMP_SfM}, a commercial photogrammetry software.

We first estimate the global camera poses of the captured images and a georeferenced sparse point cloud of the scene using a standard Structure-from-Motion (SfM) process (referred as "alignment" in RealityCapture). Georeferencing is achieved 
using Ground Control Points (GCPs) and RealityCapture's GCP annotation tool. 
Taking into account the drone-based acquisition described above, the GCP annotation, and the further processing of the data, we measure a total georeferencing root mean square error (RMSE) of \SI{5.45}{\cm} horizontally and \SI{11.60}{\cm} vertically. This implies that the data is scaled to real world units, meters in our case. Note that georeferencing is not a priority for semantic segmentation task apart from scaling the data to the real world units. Therefore, we provide all point coordinates as scaled and, for convenience, as zero-centered per tile in our dataset.

\begin{figure*}[tbp!]
	\centering
	\begin{minipage}{0.32\textwidth}
	\centering
	  \subcaptionbox{\label{fig:Zurich_center}}
	  {\includegraphics[width=\linewidth]{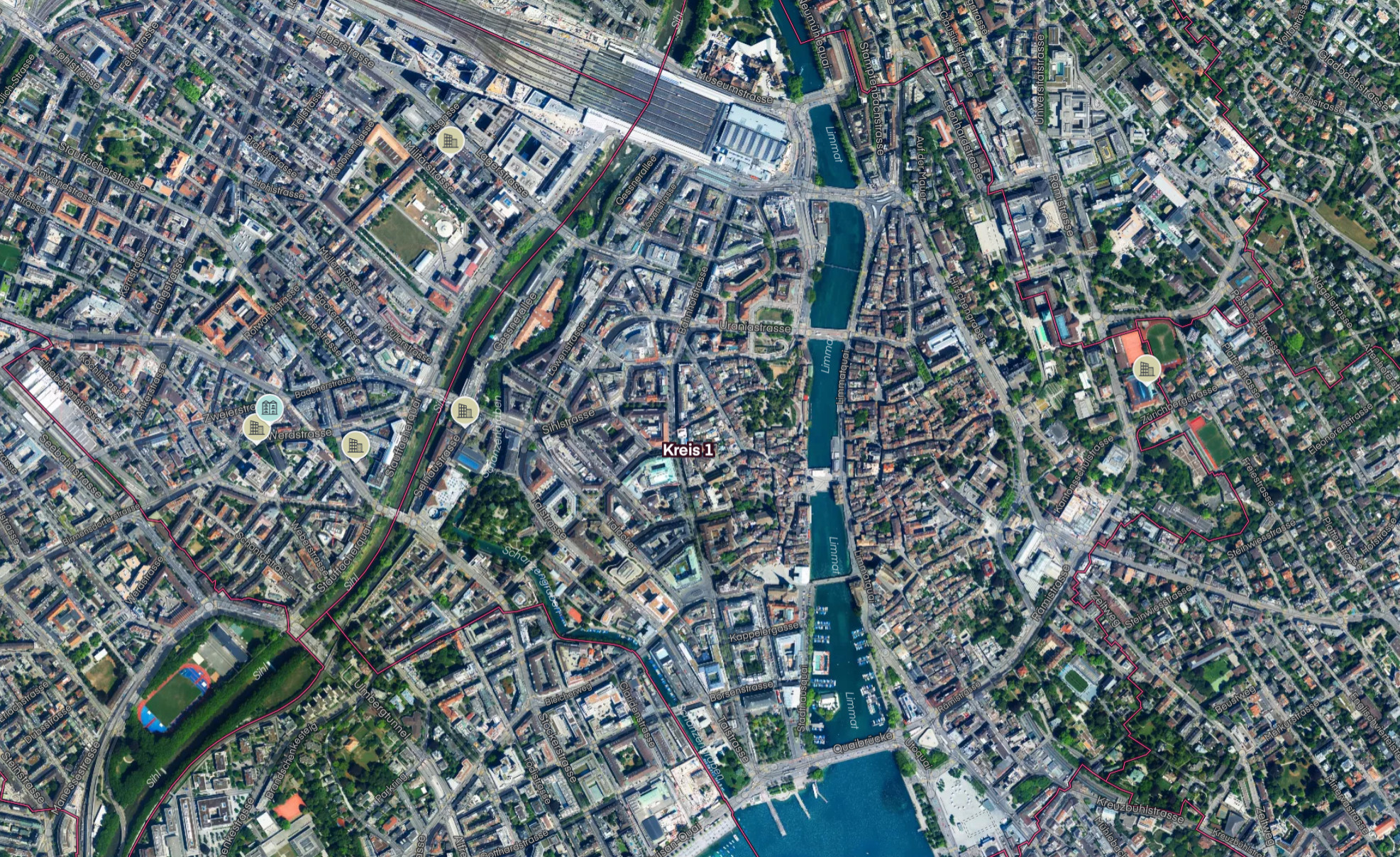}
	  }
	\end{minipage}
	\vspace*{-3mm}
	\begin{minipage}{0.325\textwidth}
	\centering
	  \subcaptionbox{\label{fig:Zug_Cham}}
	  {\includegraphics[width=\linewidth]{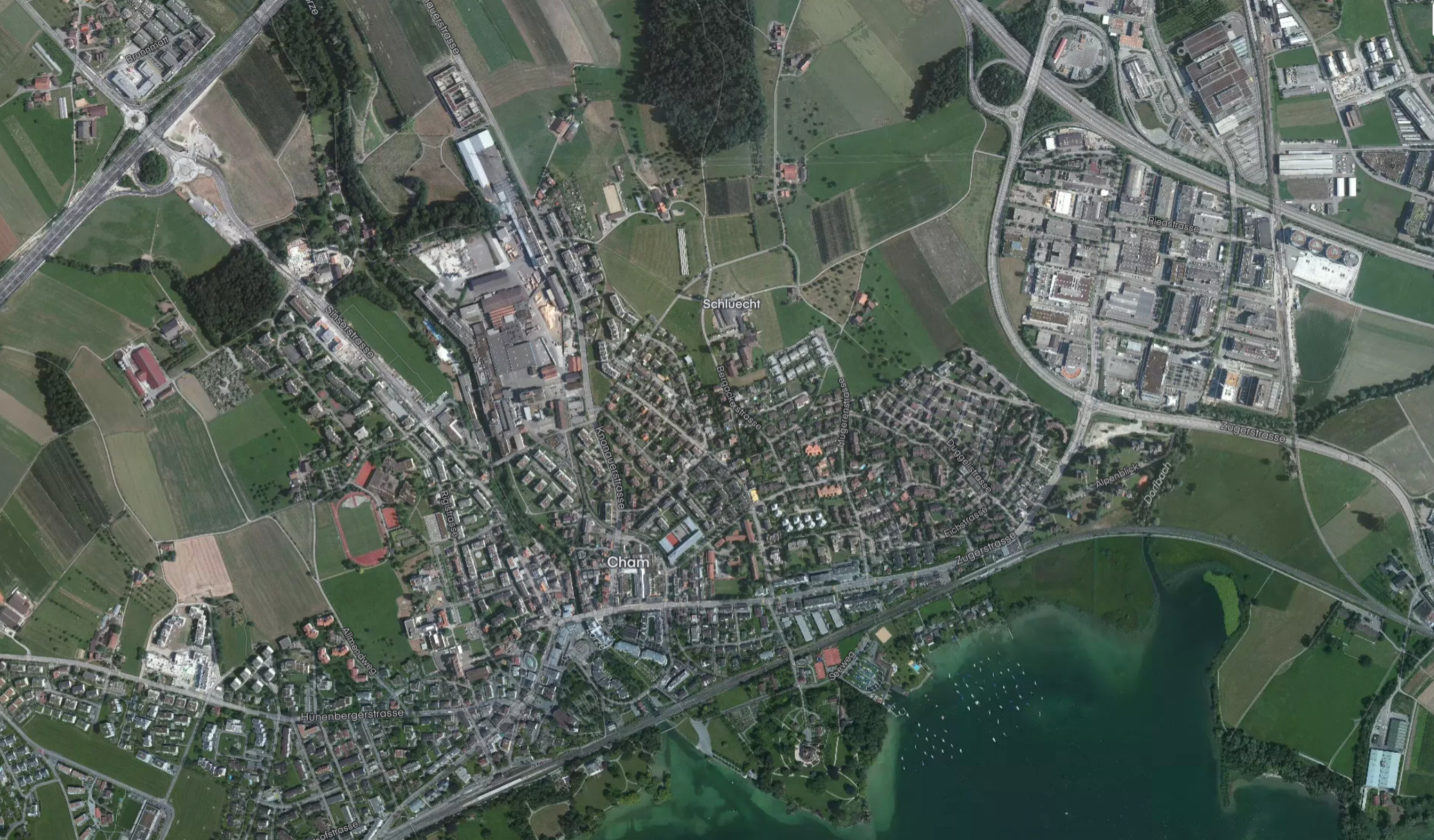} 
	 }
	\end{minipage} 
	\begin{minipage}{0.325\textwidth}
	\centering
	  \subcaptionbox{\label{fig:Davos}}
	  {\includegraphics[width=\linewidth]{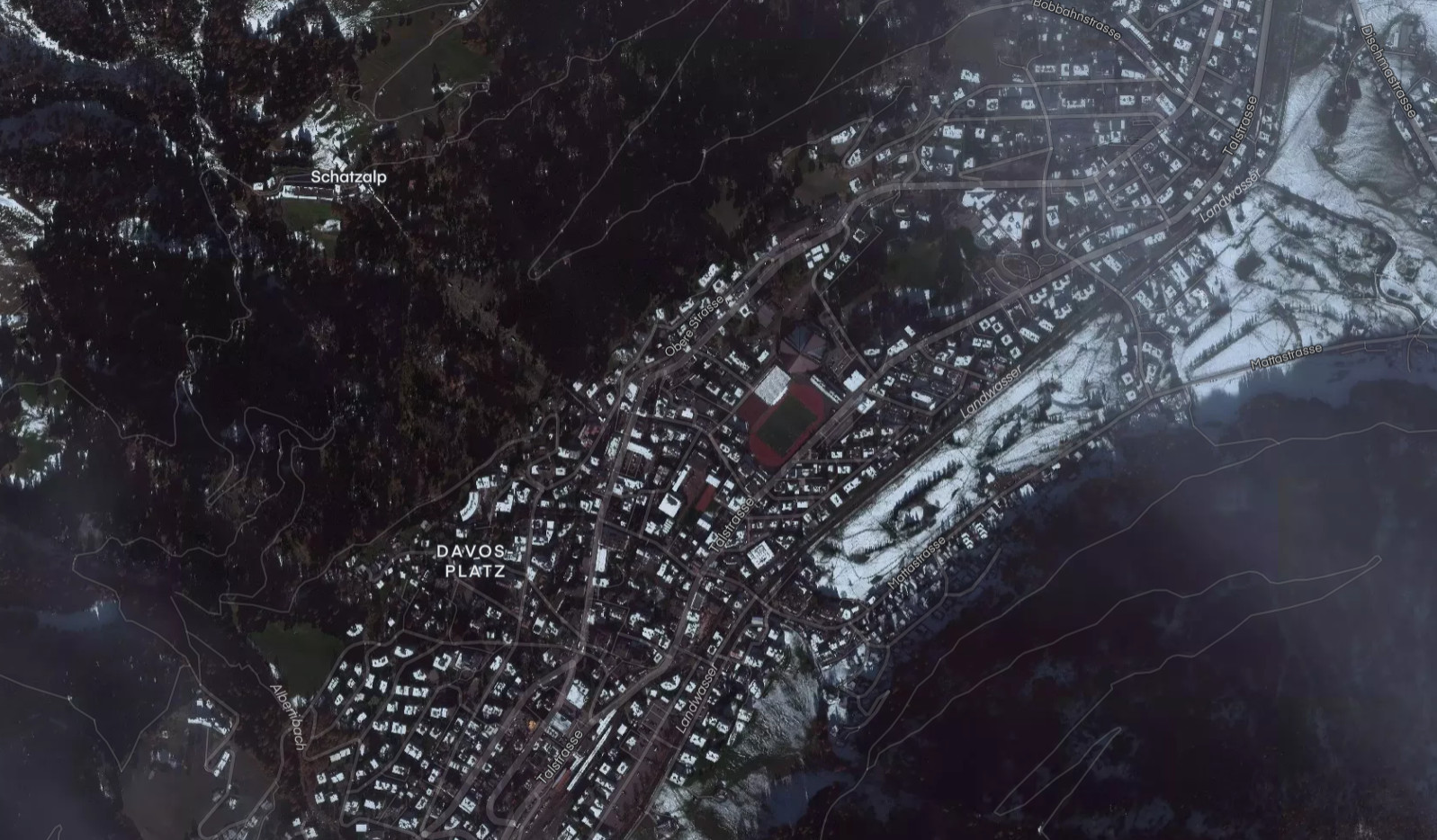}
	  }
	\end{minipage}
	\caption{A map of (a) Switzerland in Europe, (b) highly-populated urban Zurich center, (c) rural and industrial Zug-Cham region, and (d) mountainous Davos region.}
	\label{fig:cities_map}
	\vspace*{-3mm}
\end{figure*}

\begin{figure*}[tbp!]
	\centering
	\begin{minipage}{0.24\textwidth}
	\centering
	  \subcaptionbox{\label{fig:1_Davos_1M_mesh}}
	  {\includegraphics[width=\linewidth]{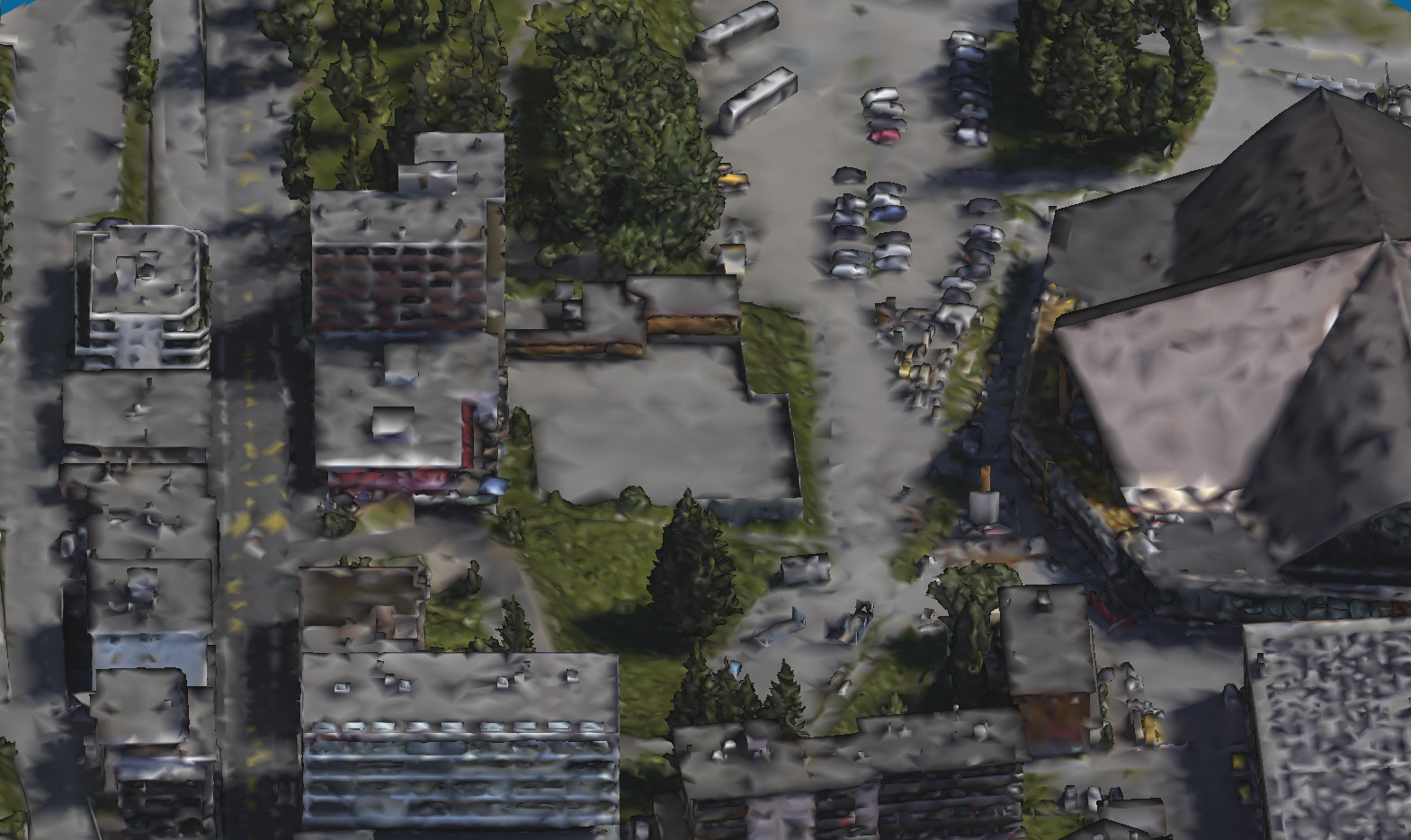}}
	\end{minipage} 
	\begin{minipage}{0.24\textwidth}
	\centering
	  \subcaptionbox{\label{fig:1_Davos_normal_pc}}
	  {\includegraphics[width=\linewidth]{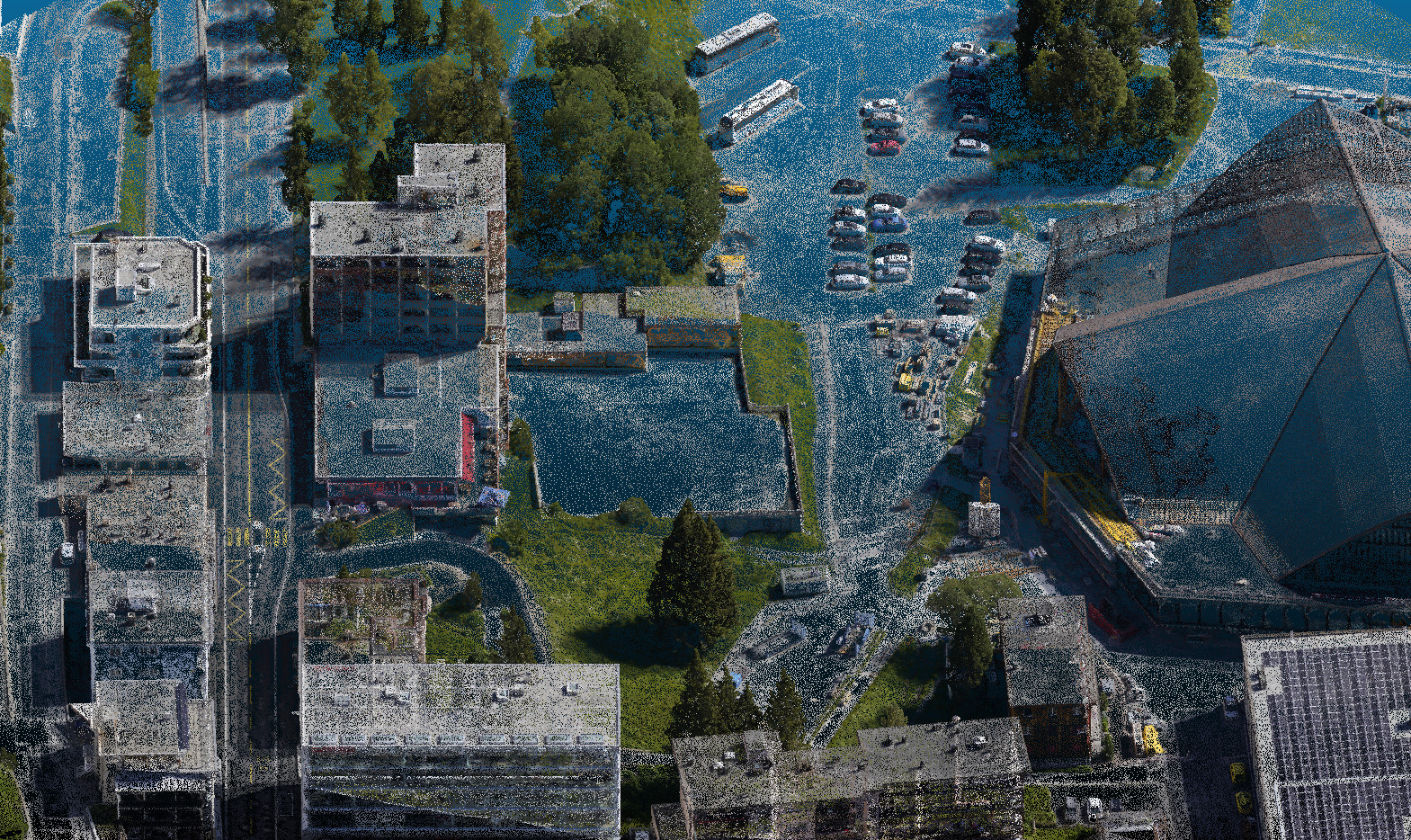}}
	\end{minipage}
	\vspace*{-3mm}
	\begin{minipage}{0.24\textwidth}
	\centering
	  \subcaptionbox{\label{fig:1_Davos_normal_pc_gt}}
	  {\includegraphics[width=\linewidth]{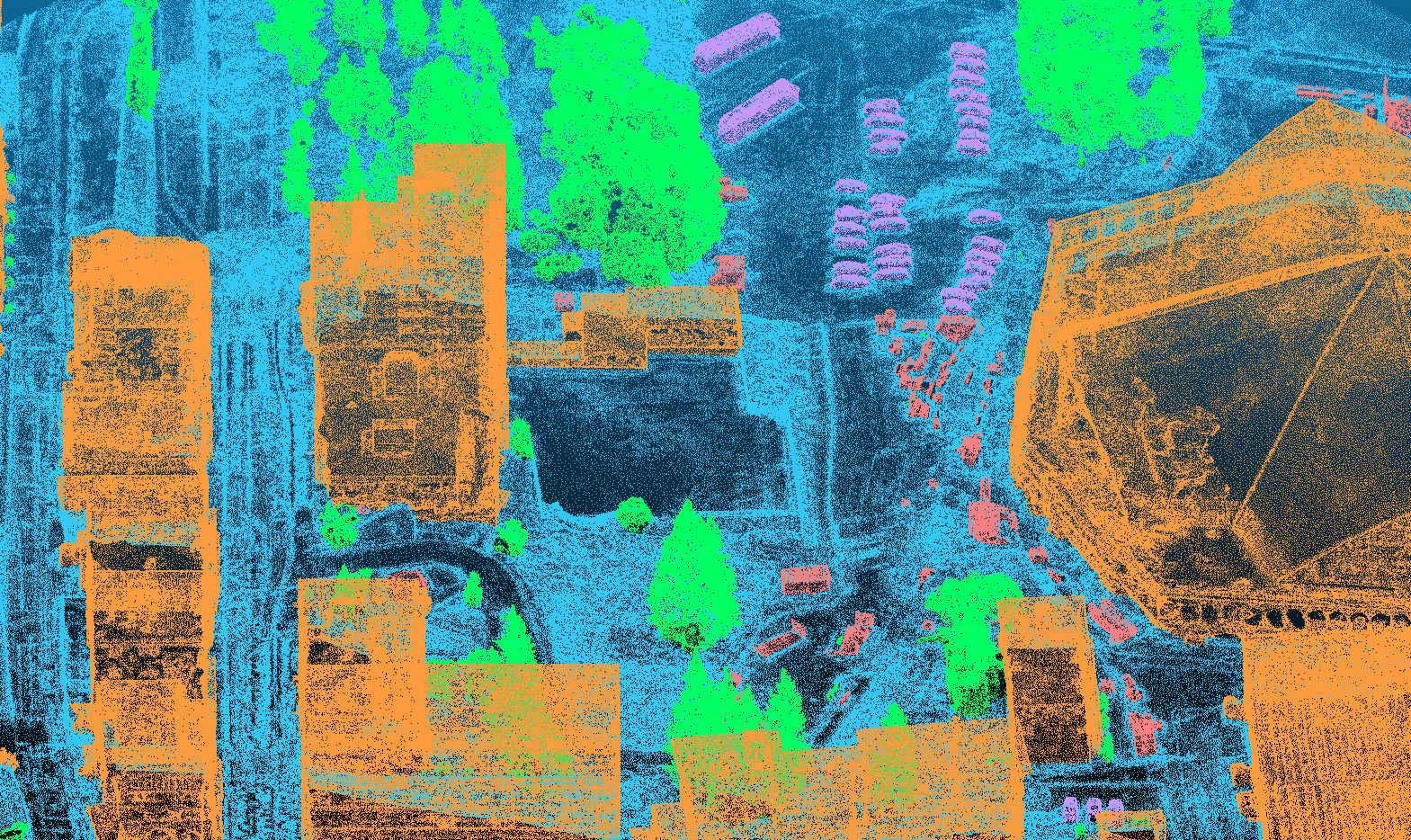}}
	\end{minipage} 
	\begin{minipage}{0.24\textwidth}
	\centering
	  \subcaptionbox{\label{fig:1_Davos_dense_pc}}
      {\includegraphics[width=\linewidth]{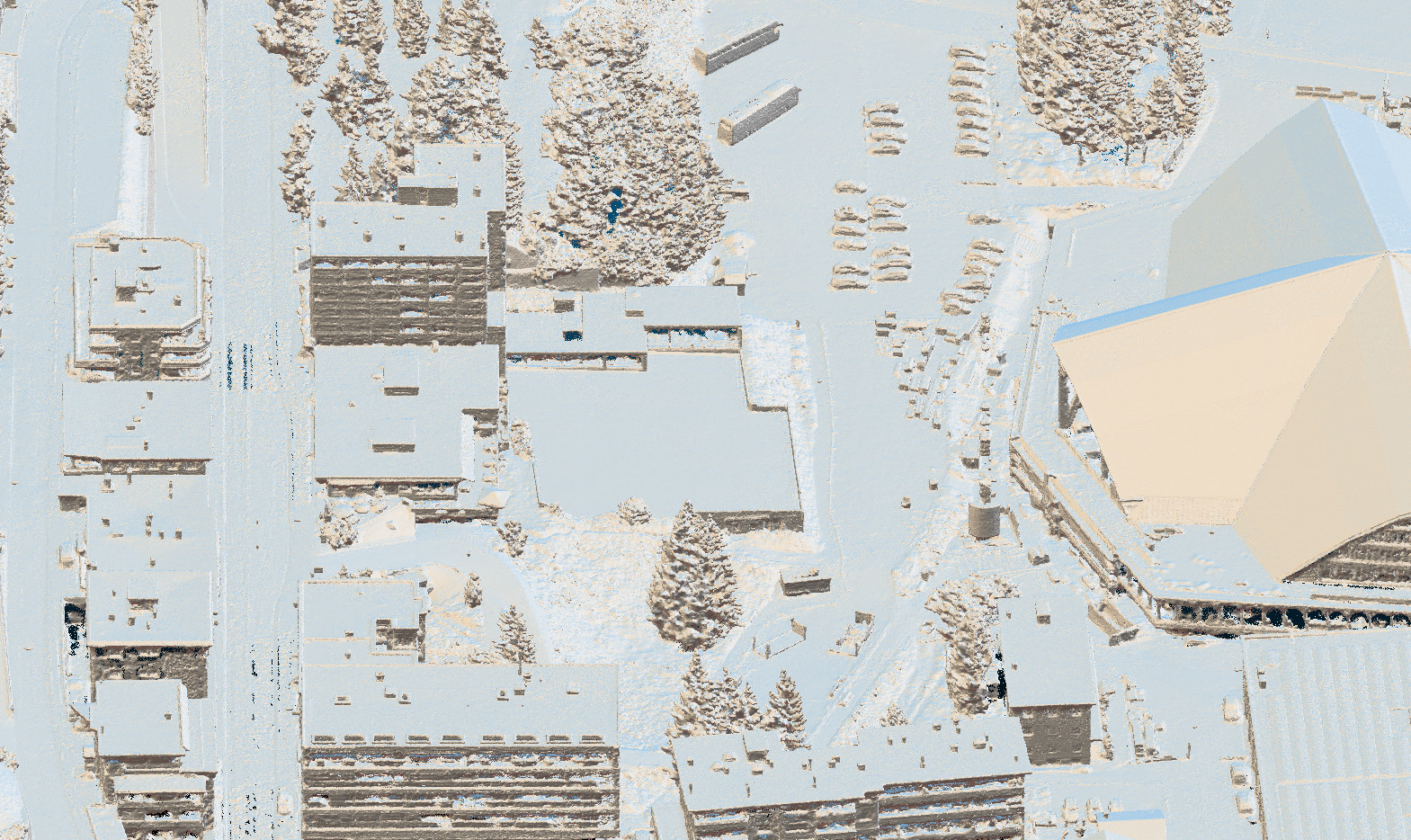}}
	\end{minipage}
	\caption{(a) A part of the simplified mesh with 1M polygons, (b) the point cloud with 15M vertices, (c) the semantic groundtruth, and (d) the dense pointcloud with around 220M points for PC 1 (Davos 16\_34318\_-22950).}
	\label{fig:1_Davos_pc_density}
	\vspace*{-3mm}
\end{figure*}


Once the data is aligned and georeferenced, we reconstruct a dense mesh constrained to the geographic region of the tile only.
At that point, the raw mesh obtained can contain up to half a billion polygons for a single tile.
In order to get a mesh with a more manageable size, we simplify it to a maximum number of 30 million polygons, i.e. approximately 15 million vertices, with Reality Capture's adaptive simplification process and texture it using the captured drone images.
The output point cloud used for segmentation is composed of the vertices of such a mesh; the RGB color of each point is sampled from the mesh texture.

\subsection{Manual Segmentation}
Figure~\ref{fig:data_pipeline} illustrates the steps after the reconstruction of a 3D model to obtain a segmented pointcloud. Our pointclouds are segmented manually into the five semantic categories as described below (terrain, construction, vegetation, vehicles and urban assets).  

3D artists complete this task using off-the-shelf 3D modeling software (such as Blender~\cite{blender}); to make the process manageable, they work on each tile individually, and operate on a 1-million polygon mesh further simplified from the initial mesh. It takes between six to twelve hours for a 3D artist to manually segment each tile.


Labels are then transferred from the simplified mesh to the output point cloud. The label of each point in the output point cloud is assigned by finding the nearest neighbor in the segmented mesh. We used an adaptive distance threshold to avoid matching outlier points.
We found that this method gives satisfying results for the final segmentation of the point cloud while keeping the amount of manual work needed at a manageable level.

\subsection{Dataset details}
\label{sec:data_details}
The dataset represents sixteen tiles acquired from three cities in Switzerland (see Figure~\ref{fig:cities_map}): six tiles from Zurich, five tiles from Zug and five tiles from Davos.


For each tile, the dataset contains pointclouds at three resolutions, i.e. approximately $500K$, $15M$, and $225M$ points per tile as shown in Figure~\ref{fig:1_Davos_pc_density}.  Both $500K$ and $15M$ point density pointclouds have x,y,z, and RGB color features. For the highest density, we have only x,y,z coordinates. 
In the rest of this paper, we only consider the 15M point density.

\paragraph{Classes and class distribution}
As our dataset is focused on urban areas in Switzerland, it comprises of a large amount of terrain, building, and medium or high vegetation. Even though many objects of other categories (such as vehicles or urban assets) are present in our dataset, they amount to a relatively small portion of the points because each object is relatively small. For that reason, we divide our semantic labeling to only five main categories: 1) terrain (including natural terrain, e.g. grass or soil, impervious terrain, e.g. road or sidewalk, and water areas, e.g. river or lake); 2) building; 3) urban asset (including traffic light, pole, crane, public transportation stop, trash bin, etc.); 4) vegetation (tree or bush); and 5) vehicle (car, bike, scooter, etc.). The total number of points per category can be found in Table \ref{tbl:n_points_per_class} as well as the mean and standard deviation of number of points among tiles.

\begin{table}[bp!]
\caption{Number of points per class}
\centering
\footnotesize
\begin{tabular}{cccc}
	\multicolumn{1}{c}{}                                           & \multicolumn{3}{c}{\textbf{Number of Points}}       \\ \cline{2-4} 
	\multicolumn{1}{c}{\textbf{Category}}                          & \textbf{Mean} & \textbf{Std. dev.} & \textbf{Total} \\ \hline
	Terrain                                               & 4,030,709     & 1,731,832.3        & 64,491,349     \\ \hline
	Construction                                          & 6,509,061     & 3,111,158.9        & 104,144,973    \\ \hline
	Urban asset & 167,595       & 121,861.0          & 2,681,512      \\ \hline
	Vegetation                                           & 3,282,801     & 1,644,790.8        & 52,524,820     \\ \hline
	Vehicle                                               & 170,662       & 84,486.5           & 2,730,595      \\ \hline
\end{tabular}
\label{tbl:n_points_per_class}
\end{table}


\section{Semantic Segmentation}
\label{sec:method}
To exemplify the usage of the proposed dataset for training and evaluating semantic segmentation models, and to provide baseline performance metrics, we report experiments using PointNet++~\cite{qi2017pointnet++}, a well-established point cloud segmentation approach.

\subsection{PointNet++}
\label{sec:model}

PointNet++~\cite{qi2017pointnet++} is a deep learning model built upon the PointNet~\cite{qi2017pointnet} model. 
In the PointNet++ architecture, PointNet module is used as a local feature encoder and applied in a nested fashion to learn hierarchical features. 
Moreover, PointNet++ uses farthest point sampling to cover more representative points during sampling. 

We adopt an existing implementation~\cite{pytorchpointnet2} of PointNet++ developed using PyTorch~\cite{pytorch_NIPS2019},  PyTorch-Lightning~\cite{falcon2019pytorch} and Hydra~\cite{Yadan2019Hydra}.
For a given instance, the input of the model is a $N \times 6$ matrix, each row containing the $x,y,z$ coordinates and $R,G,B$ color of one of $N$ input points. The output of the model is a matrix of $N \times K$ prediction probabilities, where $K = 5$ is the number of classes. Because the model is designed to handle input point clouds up to a few thousand points ($N = 8192$ in our reference implementation), it cannot be directly applied to our large outdoor datasets; therefore, we implemented the following data pipeline.  First, we partition the input data into columns with a base of $10m\times10m$ and infinite height. 
During training and validation, each instance is generated by picking a column, then randomly sampling (with replacement) $N$ points from the column.  A training epoch is obtained by generating one instance per column.  For every training epoch, the instances are sampled again; this yields a form of data augmentation since for each column a different subset of points is sampled in every epoch.

Once a model is trained, in order to segment a testing tile, we apply the model to every column separately, then merge the segmentation results.  To segment a column, we randomly divide the points in the column in subsets, each containing exactly $N$ points; for the last subset, in case less than $N$ points are remaining, additional points are sampled from the other subsets.  Each subset defines an instance an is segmented independently using the trained model; the results are then combined.


The model is trained by minimizing the cross-entropy loss; to deal with heavy class imbalance, following in similar works~\cite{hu2020sensaturban}, the loss is weighted differently for each class, according to inverse-square-root frequency.  A training batch is composed by 64 instances and we train for 200 epochs; we do not use early stopping but snapshot the model which yields the minimum loss on the validation set (which is defined on tiles different than training and testing tiles, see below). Other hyper-parameters are set as in \cite{pytorchpointnet2}.
The experiments are run on a NVIDIA RTX 2080Ti GPU. The longest train and test sessions are completed in less than 6 hours.



\subsection{Experimental Setup}
Our experimental setup focuses on the following research questions, that are more related to the characteristics of the data than to the capabilities of the specific model.
\begin{itemize}
    \item Which categories are more challenging to segment?
    \item How does the model generalize across cities? 
    \item How much can additional data help even if it is from a different city? 
    \item Which training strategy is better for pointcloud data: an ensemble of per-city models or a single model trained on all data from multiple cities? 
\end{itemize}

To answer the questions above, we train four models: three on data sampled from a single city (named \emph{single-city} models in the following); one on data from all three cities.  Then, we apply each model on three testing sets (disjoint from the training and validaiton sets), one per city.

\paragraph{Data Splits}
Table~\ref{tbl:data_split_details} provides the details of data splits for our experiments.

\begin{table}[tbp!]
	\footnotesize
	\centering
	\caption{The pointcloud IDs used in train, validation and test sets for different experiments.}
	\label{tbl:data_split_details}
    \begin{tabular}{ccccc}
                                                          &  \multicolumn{3}{c}{\textbf{Cross-city}} & \textbf{Full}                                                                                              \\ \cline{2-5} 
\textbf{Model} & \multicolumn{1}{c}{\textbf{\begin{tabular}[c]{@{}c@{}c@{}}M1\\ (Davos)\end{tabular}}}   & \multicolumn{1}{c}{\textbf{\begin{tabular}[c]{@{}c@{}c@{}}M2\\ (Zug)\end{tabular}}}     &  \multicolumn{1}{c}{\textbf{\begin{tabular}[c]{@{}c@{}c@{}}M3\\ (Zurich)\end{tabular}}}      &   \textbf{M4}                                                                                                                                                           \\ \cline{1-5}
\textbf{Train Set}                                                    & 1, 2, 3 & 6, 7, 8 & 11, 12, 13 & \begin{tabular}[c]{@{}c@{}}1, 2, 3, \\ 6, 7, 8, \\ 11, 12, 13\end{tabular} \\ \hline
\textbf{Val. Set}                                                     & 4       & 9       & 14                                                         & 4, 9, 14                                    \\ \hline
\textbf{Test Set}                                                     & 5, 10, 15       & 5, 10, 15      & 5, 10, 15         & 5, 10, 15                                                                                     \\ \hline
\end{tabular}
\end{table}


In particular, we consider five tiles for each of the three cities. Each tile covers approximately \SI{0.17}{\km\squared}, which yields \SI{0.855}{\km\squared} and 70 million points per city.

For each city, the five tiles are partitioned in three tiles for training, one tile for validation and one tile for testing.  Single-city models are therefore trained on three tiles and validated on one tile.  The model trained on all cities is trained on nine tiles and validated on three tiles.  Each of the four models is tested on three tiles (one per city), on which we separately compute performance metrics.  


\paragraph{Evaluation Metrics}
For a given testing tile, a model under test will produce five class probabilities (which sum to 1) for each point. The point is then assigned to the class that has the largest probability. From these data, we compute the following commonly-used metrics~\cite{armeni20163dS3DIS,semantic3d_hackel2017isprs,behley2019semantickitti} to quantify segmentation performance.
\begin{itemize}
    \item \emph{Overall Accuracy} is the fraction of the points for which the predicted class coincides with the ground truth class (also known as micro-averaged accuracy).
    \item \emph{Weighted Accuracy} is the macro-averaged accuracy 
    that is multiplied with a per-class factor. For a given class $c$, the factor is computed as the proportion of the number of class samples $N_c$ over the number of samples in the whole dataset $N$, i.e. $N_c / N$.
    \item \emph{Per-class F1 score} is the harmonic mean between per-class precision and recall.  An F1 score of 1.0 indicates an ideal classifier.
    \item \emph{Per-class Intersection over Union score (IoU):} For a given class $c$, the IoU score is computed as the ratio between: the number of points that have been classified as class $c$ AND are indeed of class $c$ (intersection); and the number of points have been classified as class $c$ OR are indeed of class $c$ (union). An IoU score of 1.0 indicates an ideal classifier.
\end{itemize}
For all per-class metrics, we also report average values among all the classes and the weighted averages. For consistency, we report all the metrics as percentage values (the ratios between 0 and 1 are scaled linearly between 0 and 100).
\section{Results and Discussion}
\label{sec:result}

\begin{table*}[tp]
\caption{Per-class performance metrics for the model trained on all cities, when evaluated on one testing tile for each city (rows). The metrics are presented as percentage values.}
\label{tab:classperformance}
\scriptsize
\begin{tabularx}{\textwidth}{lXXXXXXXXXX}
\toprule
{} & \multicolumn{2}{c}{Terrain} & \multicolumn{2}{c}{Construction} & \multicolumn{2}{c}{Urban asset} & \multicolumn{2}{c}{Vegetation} & \multicolumn{2}{c}{Vehicle} \\
\cmidrule(lr){2-3}
\cmidrule(lr){4-5}
\cmidrule(lr){6-7}
\cmidrule(lr){8-9}
\cmidrule(lr){10-11}
{} &     IoU &  F1 &          IoU &  F1 &         IoU & F1 &        IoU &  F1 &     IoU &  F1 \\
\midrule
Davos  &   66.6 & 80.0 &        69.7 & 82.1 &        4.2 & 8.0 &      80.7 & 89.3 &   13.1 & 23.2 \\
Zug    &   75.7 & 86.2 &        71.0 & 83.0 &        2.8 & 5.5 &      62.6 & 77.0 &   17.0 & 29.0 \\
Zurich &   48.3 & 65.2 &        81.2 & 89.6 &        5.0 & 9.4 &      58.3 & 73.6 &   24.0 & 38.8 \\
\midrule
Average&   63.5 & 77.1 &        74.0 & 85.0 &        4.0 & 7.6 &      67.2 & 80.0 &   18.0 & 30.3
\\
\bottomrule
\end{tabularx}
\end{table*}

\subsection{Overall Performance Metrics}
On the three testing tiles, the model trained on all cities yields an overall accuracy of 82.8\%, weighted accuracy of 87.6\%, average F1 of 56.0\%, and average IoU score of 45.3\%. 

\begin{figure*}[tbp]
	\centering
	\begin{minipage}{0.32\textwidth}
	\centering
	  \subcaptionbox{\label{fig:partial_pc5_Davos_16_34558_-23105}}
	  {\includegraphics[width=\linewidth]{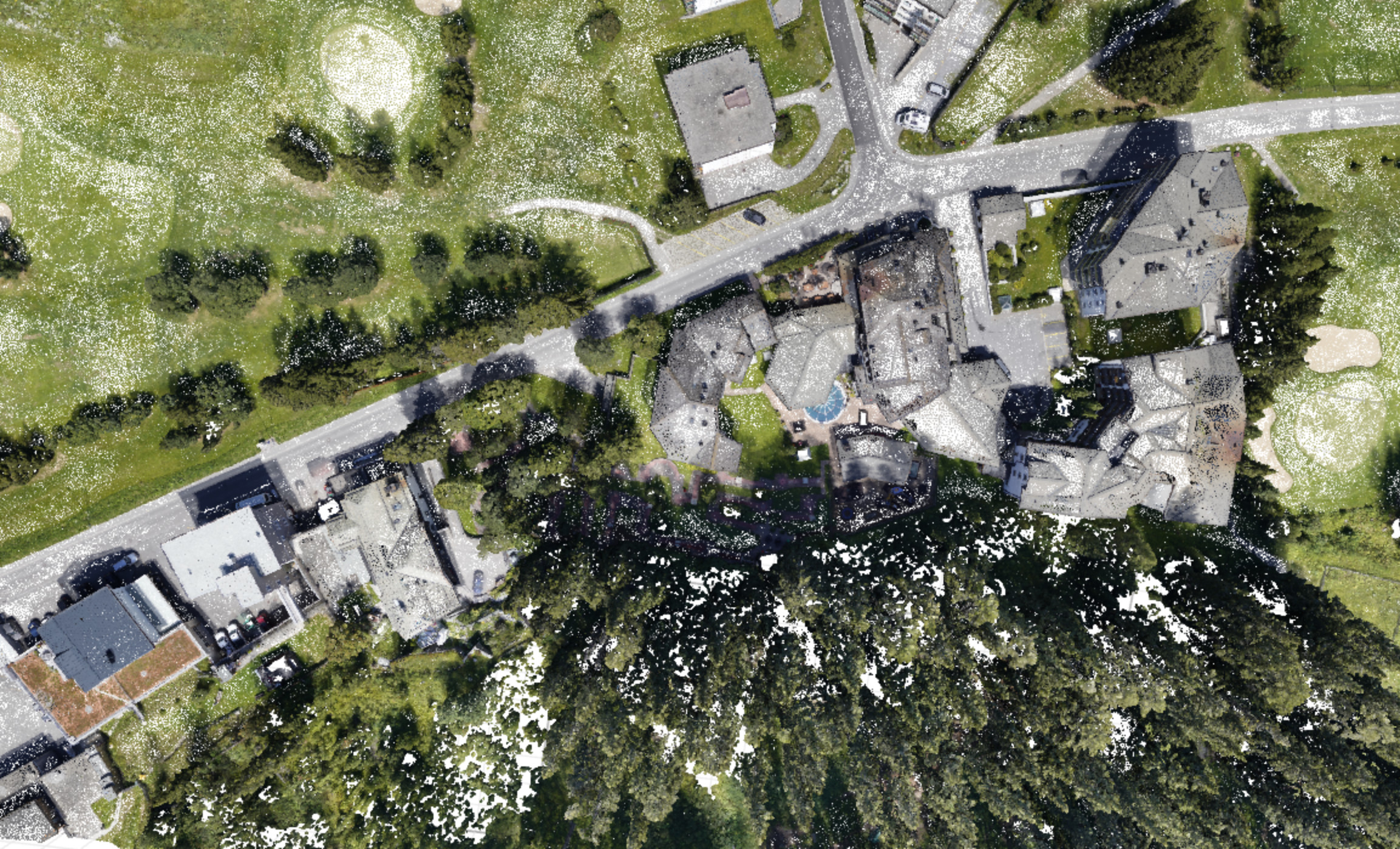}}
	\end{minipage} 
	\begin{minipage}{0.32\textwidth}
	\centering
	  \subcaptionbox{ \label{fig:partial_pc5_Davos_16_34558_-23105_predictions}}
	  {\includegraphics[width=\linewidth]{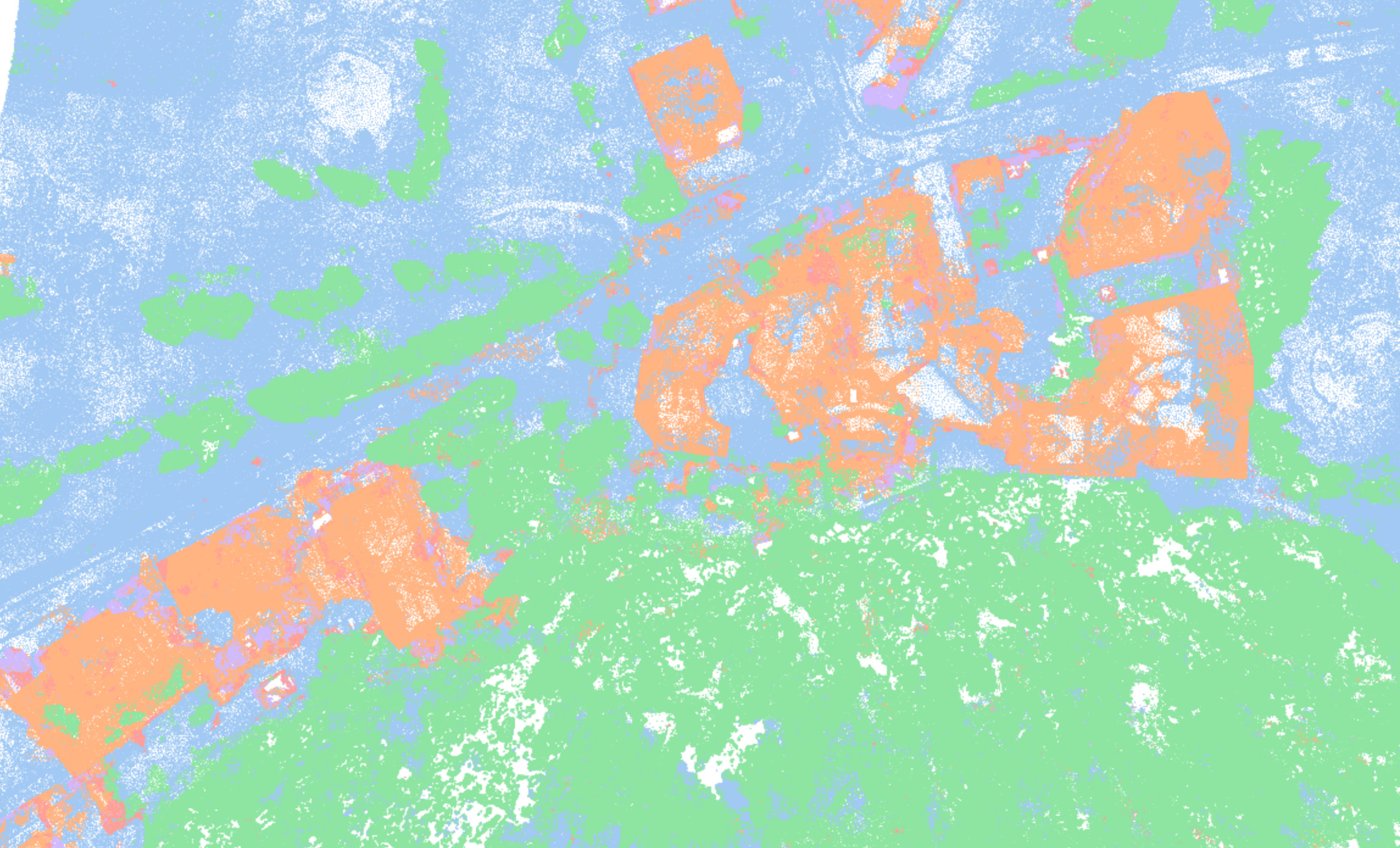}}
	\end{minipage}
	\begin{minipage}{0.32\textwidth}
	\centering
	  \subcaptionbox{ \label{fig:partial_pc5_Davos_16_34558_-23105_GT}}
	  {\includegraphics[width=\linewidth]{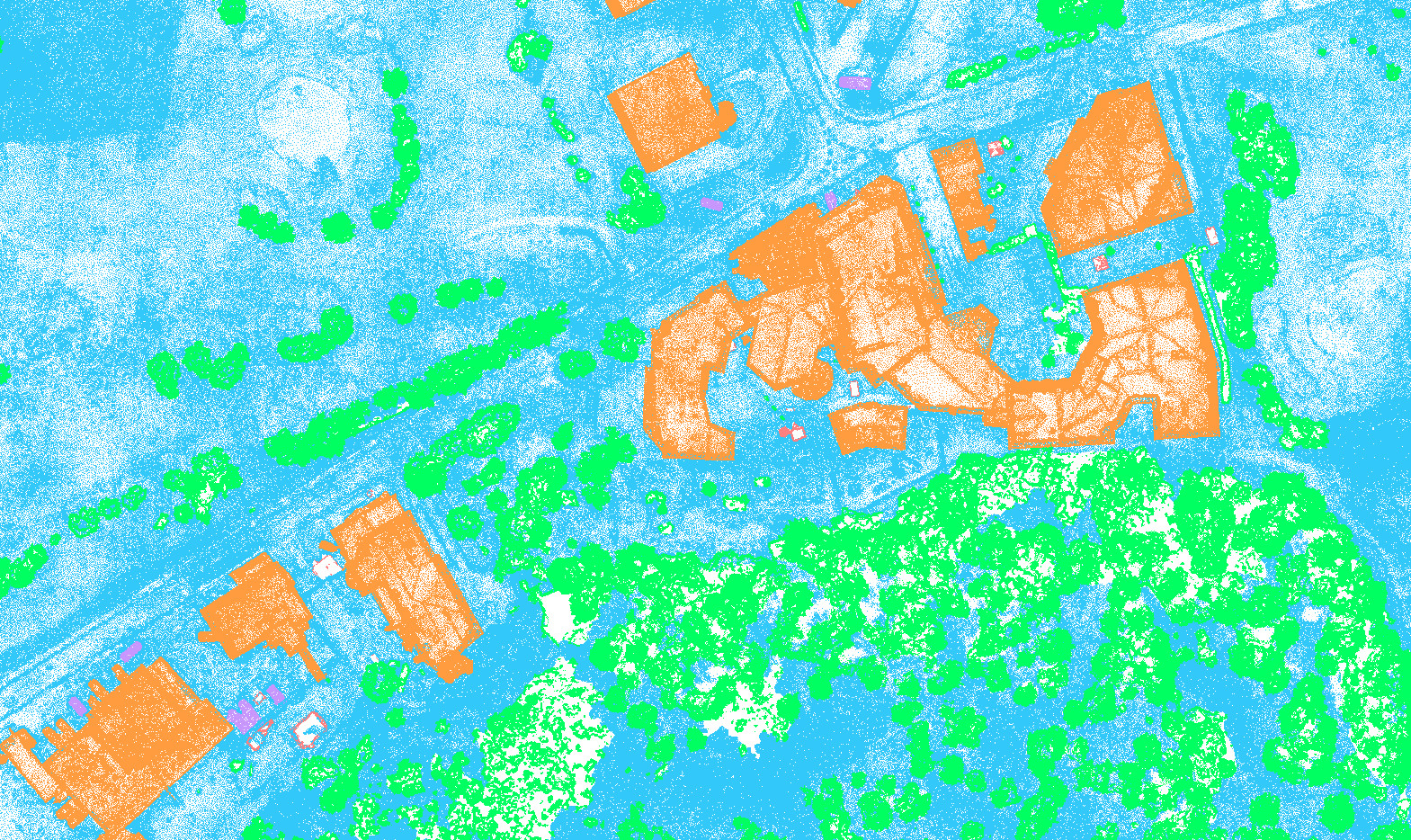}}
	\end{minipage}
	\begin{minipage}{0.31\textwidth}
	\centering
	  \subcaptionbox{\label{fig:partial_pc15_Zurich}}
	  {\includegraphics[width=\linewidth]{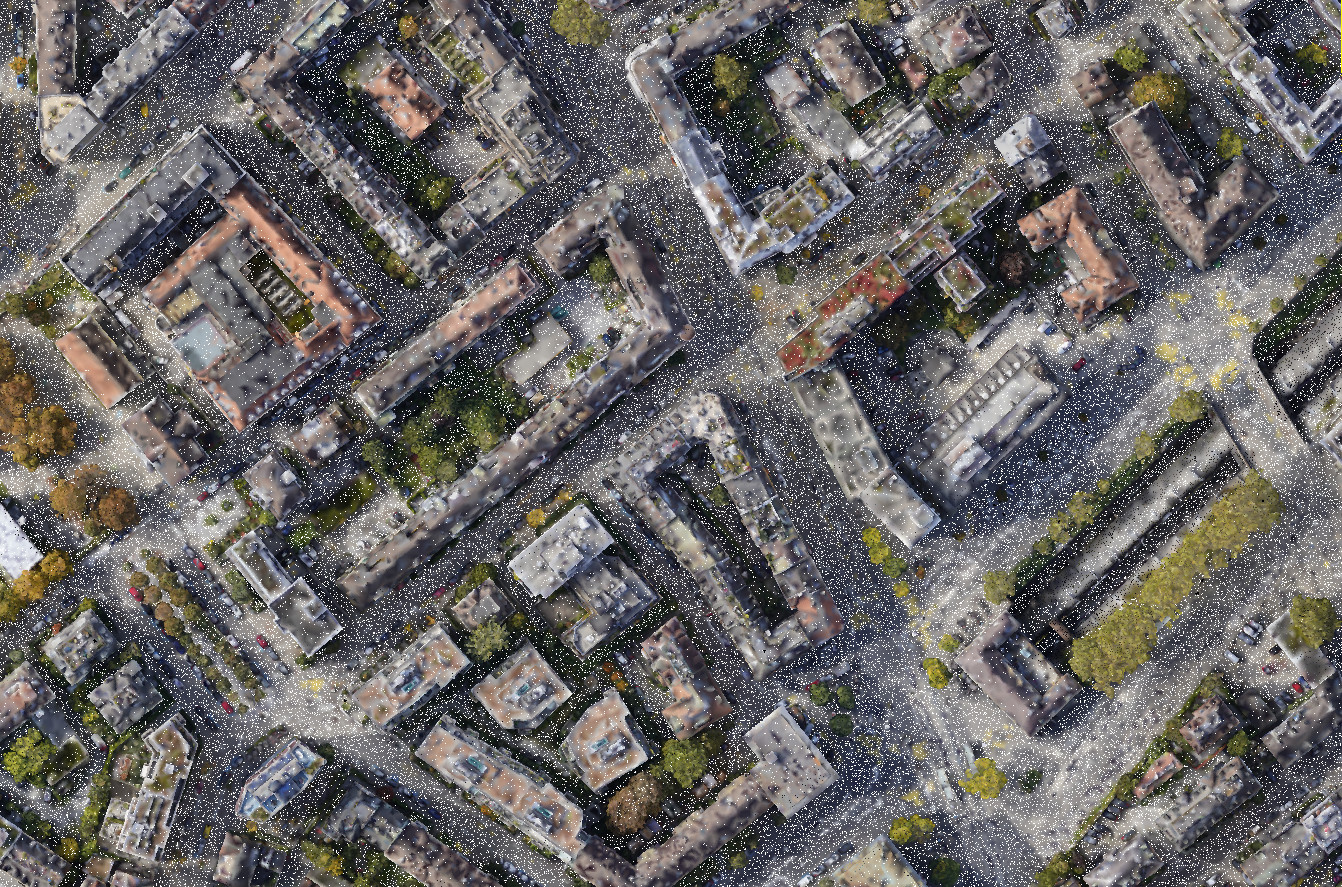}}
	\end{minipage} 
	\begin{minipage}{0.33\textwidth}
	\centering
	  \subcaptionbox{ \label{fig:partial_pc15_Zurich_predictions}}
	  {\includegraphics[width=\linewidth]{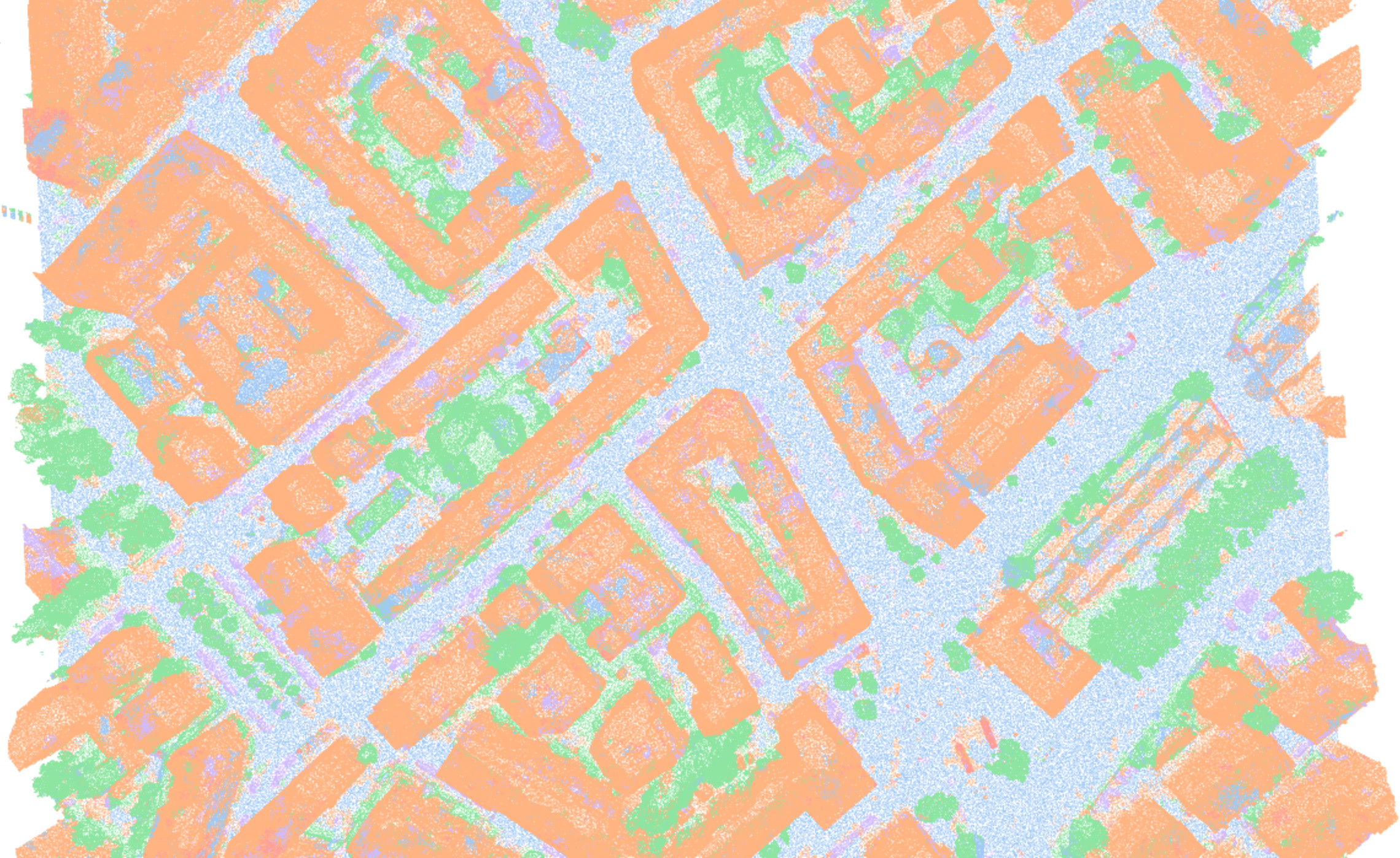}}
	\end{minipage}
	\begin{minipage}{0.32\textwidth}
	\centering
	  \subcaptionbox{ \label{fig:partial_pc15_Zurich_GT}}
	  {\includegraphics[width=\linewidth]{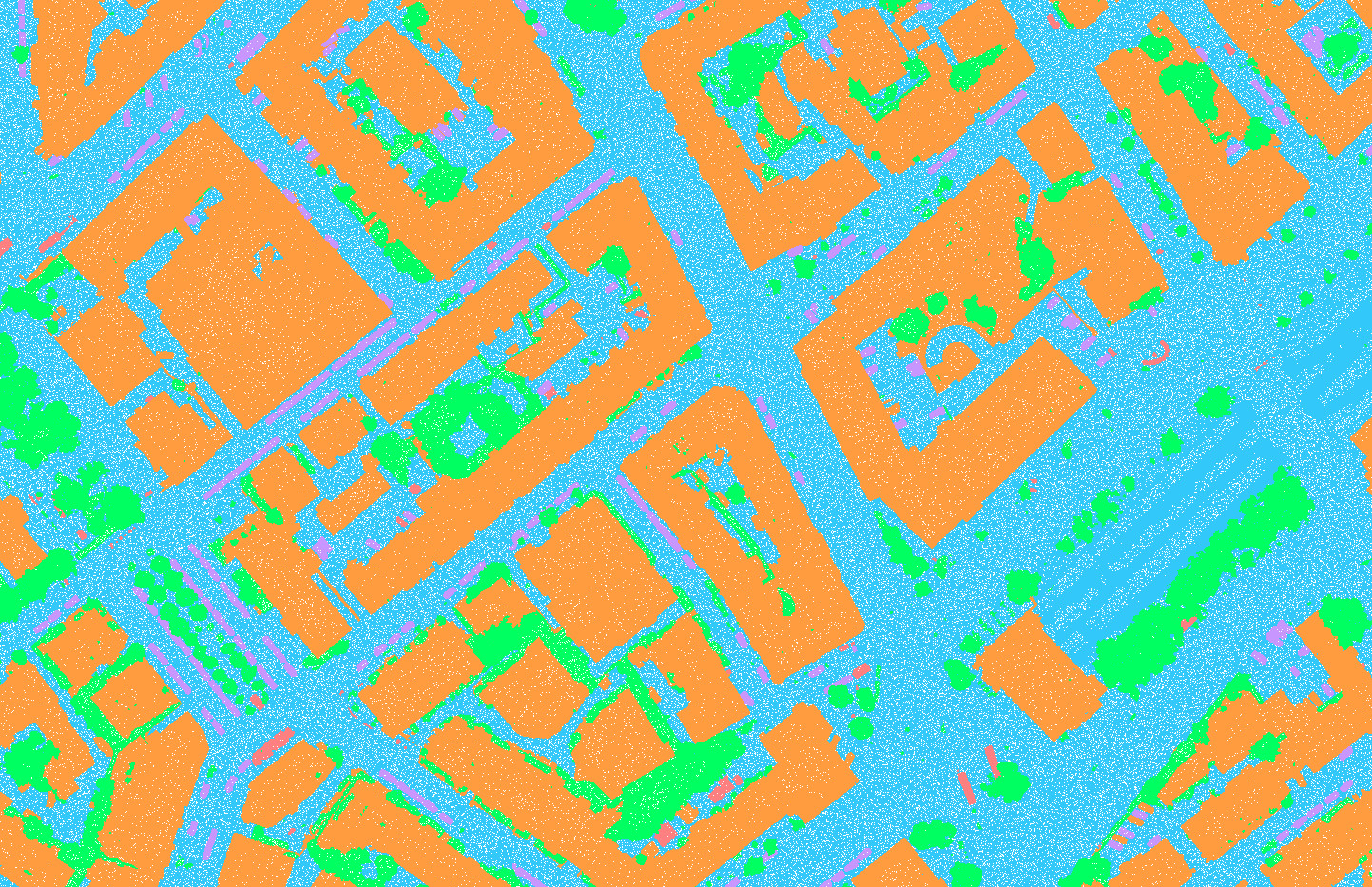}}
	\end{minipage}
	\caption{(a) The RGB visualization, (b) semantic predictions with the full model, and (c) the groundtruth of a part of PC 5 from Davos; (d) the RGB visualization, (e) semantic predictions with the full model, and (f) the groundtruth of a part of PC 15 from Zurich. Legend: blue: terrain, orange: construction, green: vegetation, purple: vehicle, pink: urban asset.}
	\label{fig:M6_prediction_pc5_pc15}
	\vspace*{-3mm}
\end{figure*}

\subsection{Per-Category Performance}
Table~\ref{tab:classperformance} reports, for each city and for each class, the performance of the model trained on data from all cities; namely, we report the per-class F1 score and IoU metrics.

\begin{figure*}[bp]
    \centering
    \begin{minipage}{0.35\textwidth}
	\centering
	  \subcaptionbox{\label{fig:M6_three_test_tiles_confusion_matrix}}
	  {\includegraphics[width=\linewidth]{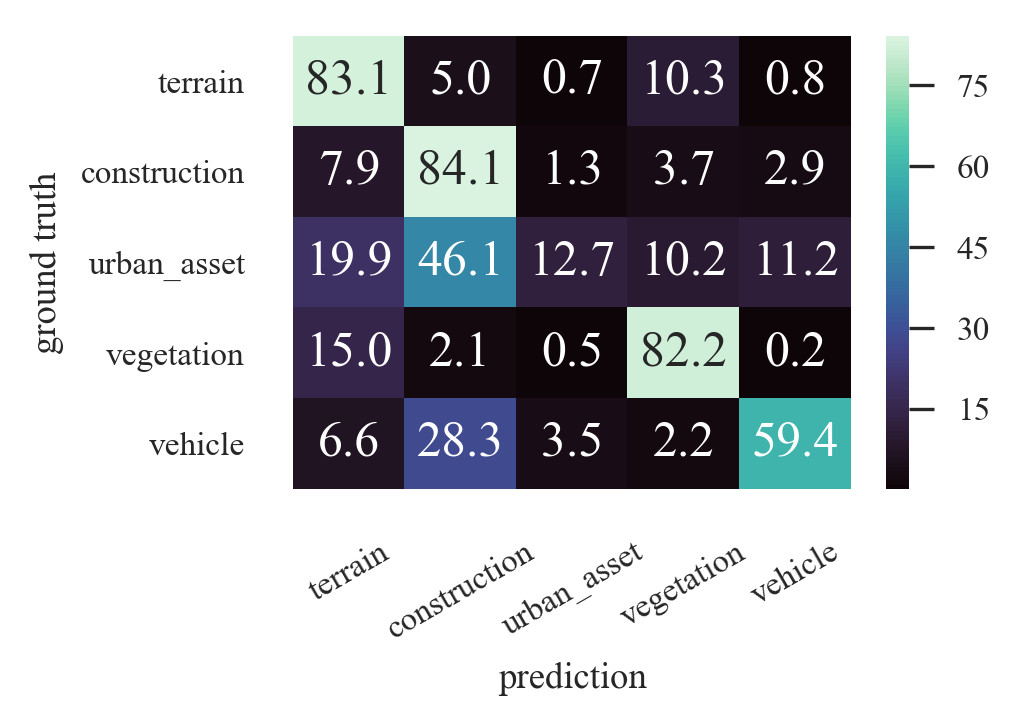}}
	\end{minipage} 
	\begin{minipage}{0.35\textwidth}
	\centering
	  \subcaptionbox{ \label{fig:M6_Zurich_pc4_test_tile_confusion_matrix}}
	  {\includegraphics[width=\linewidth]{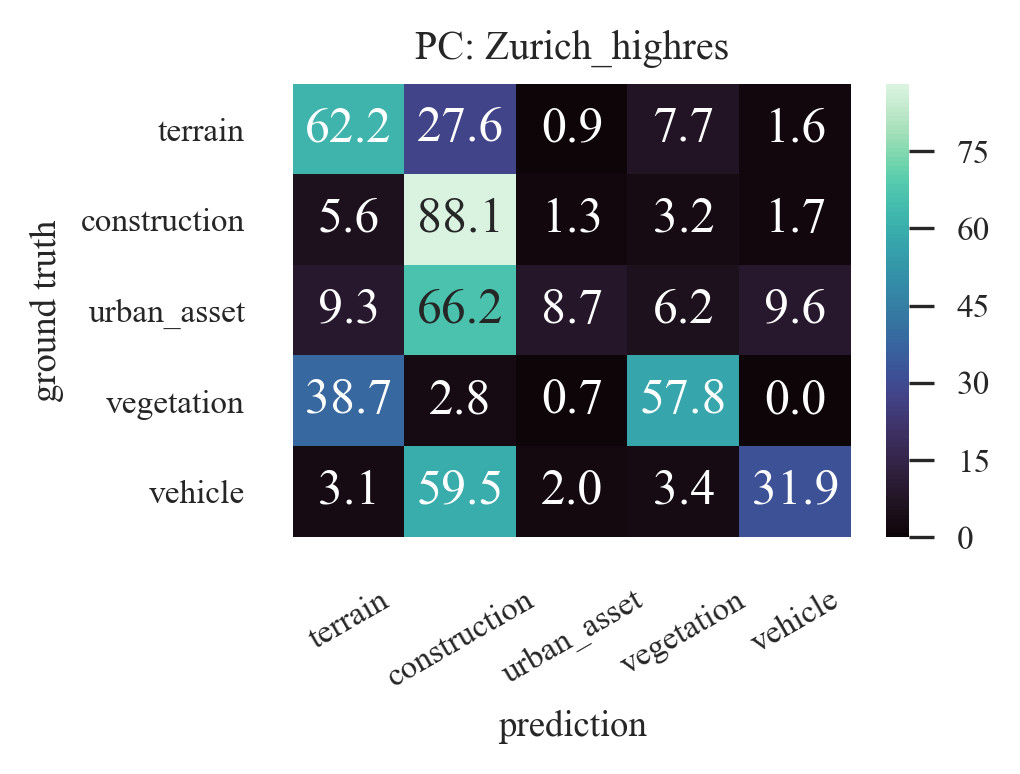}}
	\end{minipage}
    \caption{(a) The normalized confusion matrix of the model M4 (trained on all cities and evaluated on the three test tiles).  Every row reports the percentage of points with a given true class, that are classified as each of the five classes (columns). (b) The normalized confusion matrix of the model M4 on the additional urban Zurich tile (PC 16).}
    \label{fig:M6_confusion_matrix}
\end{figure*}

\begin{figure*}[tbp]
	\centering
	\includegraphics[width=0.7\linewidth]{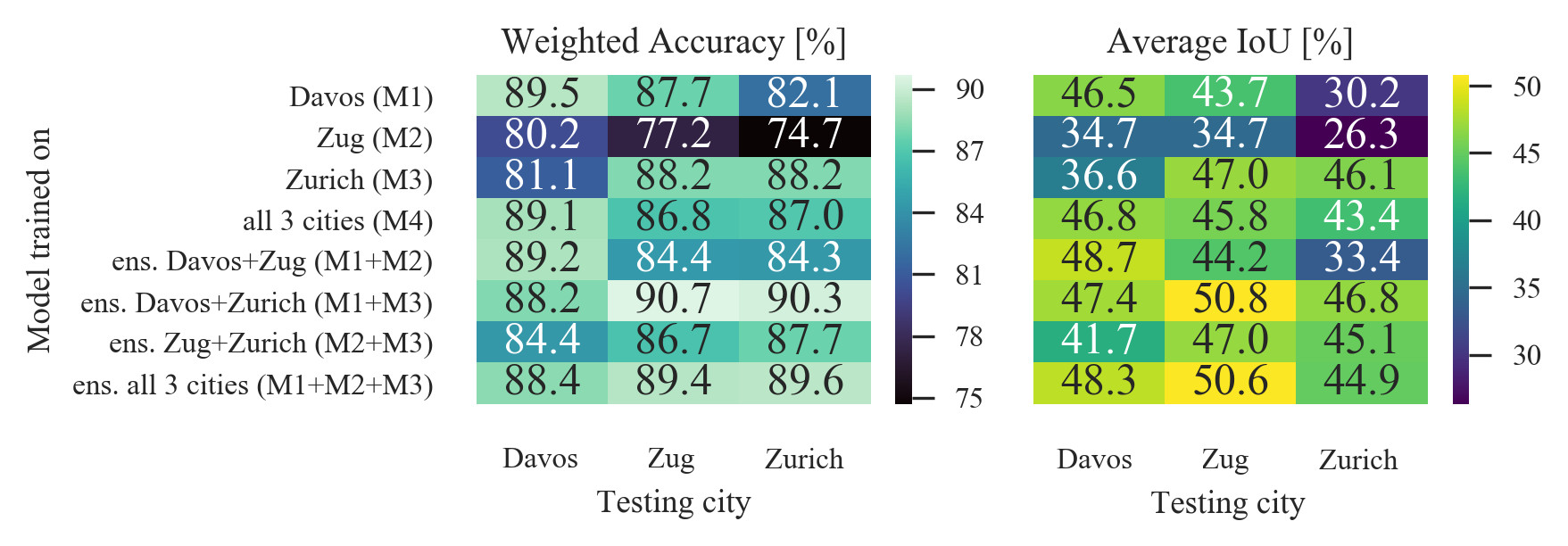}
	\caption{The weighted accuracy (left) and the average IoU (right) for different models and model ensembles (rows) evaluated on different cities (columns).  We consider: three models trained on data from each city independently; one model trained on data from all three cities; three ensembles of pairs of single-city models; one ensemble of three single-city models.}
	\label{fig:performance_table}
\end{figure*}

We observe that ``urban asset'' and ``vehicle'' classes are harder to segment compared to other classes; this is expected due to their small size, and widely variable characteristics in terms of shape and color. IoU metrics are particularly penalized, due to the small size of each object.

Figure~\ref{fig:M6_prediction_pc5_pc15} illustrates qualitative results of the full model for two test regions (a rural region from PC 5 in Davos and an urban region from PC 15 in Zurich). As illustrated in the confusion matrix in Figure~\ref{fig:M6_three_test_tiles_confusion_matrix}, one can observe the following confusion cases among categories: 1) urban asset and other categories (especially construction), 2) vegetation and terrain, and 3) vehicle and construction categories. 
The confusion matrix of an additional urban test tile from Zurich (PC 16) is shown in Figure~\ref{fig:M6_Zurich_pc4_test_tile_confusion_matrix}. This confusion matrix shows similar trends the overall confusion matrix (average of the three test tiles). However, one can notice the increase in the confusion trends and additional confusion between terrain and construction categories. This pointcloud exhibits particular urban characteristics such as a bridge, entrance to an underground parking lot, a botanical garden on a hill, and glass or plants/moss/soil covered rooftops. We consider that the lower amount of terrain in this urban setting also makes the confusion noticeable.


We hypothesize that a data pipeline that emphasizes the relative height information and favors the small categories in a stronger fashion than our current setting (e.g. a cube-based sampling rather than column-based sampling) and a stronger model than PointNet++ might help decreasing these confusion cases. As our goal is to report a baseline model on our novel dataset, we keep these model explorations for future work.
For simplicity, we report and discuss only on the three testing tiles (one per city) further in this section.

 \subsection{Model Generalization across Cities} 
We analyze the model generalization in a cross-city experiment setting.

Figure~\ref{fig:performance_table} reports performance metrics for different models, evaluated separately for each of the three cities.  
As seen in the first two rows and last column, the performance of the M1 and M2 models, which were trained on rural or industrial areas in Davos and Zug, decreases when they are tested on the urban Zurich test tile. Similarly, the Zurich model (M3), which is trained with the urban pointclouds, i.e. high-rise large buildings, performs worse on the rural Davos test tile (see third row, first column) than the other test tiles. This trend is observed further in the ensemble results. This point emphasizes the importance of area characteristics while learning semantics. 
 
\begin{figure*}[tbp]
    \centering
    \includegraphics[width=0.75\textwidth]{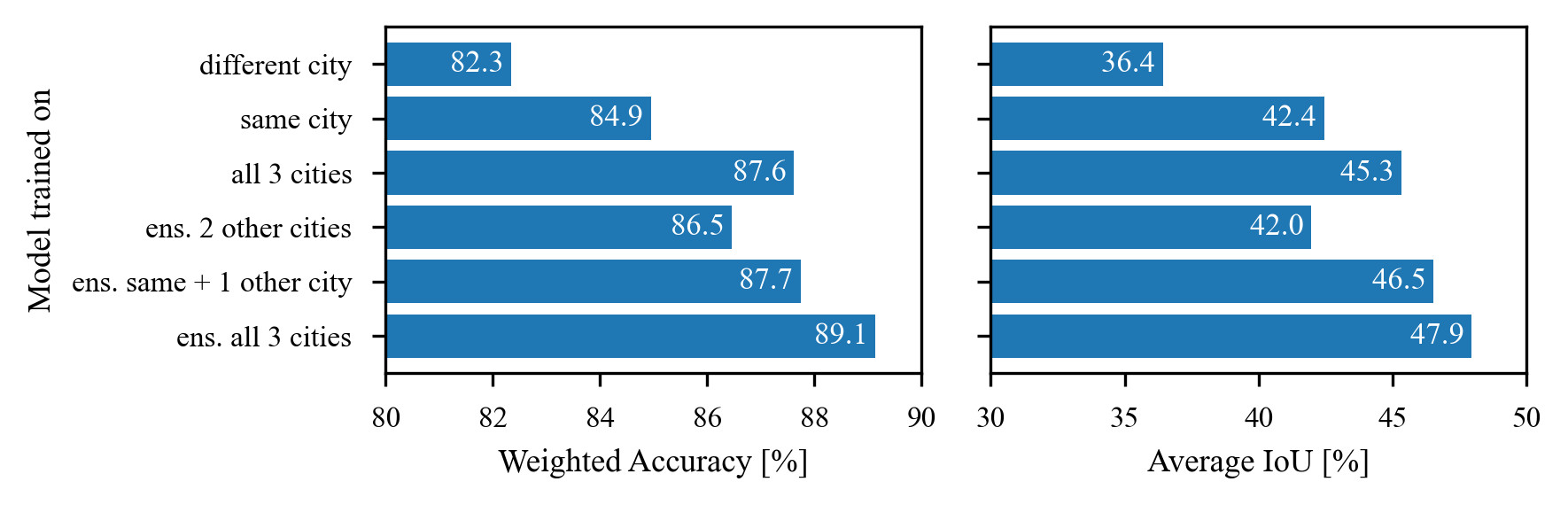}
    \caption{The weighted accuracy (left) and average IoU (right) for the models and ensembles (bars) trained on the same or different cities.}
    \label{fig:performance_bar}
\end{figure*}

Figure~\ref{fig:performance_bar} summarizes the same data by reporting the average performance of models depending on whether they are trained on the same or different cities.  For example, the performance of models trained on a city different than the testing city (first bar) is computed as the average of six performance values: two values predicted by the Davos model (M1) on the Zug and Zurich test tiles; two values predicted by the Zug model (M2) on the Davos and Zurich test tiles; two values predicted by the Zurich model (M3) on Davos and Zug test tiles.

Comparing the first and second bar of Figure~\ref{fig:performance_bar}, we observe that the models trained on data from the same city have significantly better performance (average weighted accuracy 84.9\%) than the models trained on data from a different city (average weighted accuracy 82.3\%), despite the fact that the amount of training data is the same in different cities, and that areas used for training are always disjoint from areas used for testing.

\subsection{Impact of Data Scale}
Comparing the second and third bar of Figure~\ref{fig:performance_bar}, we observe that the performance of the model M4 trained on data from all three cities (average weighted accuracy 87.6\%) is better than the performance of the model trained just on data from the same city (average weighted accuracy 84.9\%). This quantifies the impact of tripling the amount of training data, even though the additional data comes from two different cities.

\subsection{Model Ensembling}
We consider an alternative approach to training a single model on three cities; instead, we consider the three single-city models (M1, M2, and M3), apply each model independently to each test tile, then average their predictions; in particular, for each given point in the test tile, we obtain three class probability vectors as outputs of each of the three models; we compute the element-by-element average of the three vectors, which yields a single class probability vector whose 5 elements also sum to 1. This approach is known as model ensembling~\cite{hashem1997,clemen1989combining} and is used frequently in machine learning.

Comparing the third and sixth bar of Figure~\ref{fig:performance_bar}, we observe that the ensemble of the three single-city models outperforms the single model trained on the three cities.  The ensembling approach is appealing, since training each model on a single-city dataset is simple and flexible: by averaging their results, we minimize the consequences of overfitting and more generally counteract model variance; on the other hand, the computational cost for inference is tripled, as three models have to be evaluated for each input.

Model ensembling experiments also allow us to quantify the performance gains from  acquiring additional training data; in particular, by comparing the first and fourth bar of Figure~\ref{fig:performance_bar}, we can observe the benefits of building an ensemble by adding a model trained on a different city; the sixth bar shows decreasing returns when adding a third model to the ensemble, even if it is trained on the same city used for evaluation.
\section{Conclusion}
\label{sec:conclusion}
This paper introduces a novel urban pointcloud dataset with pointwise semantic groundtruth. The dataset is constructed via photogrammetry on UAV-acquired high-resolution images of three Swiss cities.
The dataset reports three pointcloud densities: a simplified sparse pointcloud with RGB colors and semantic labels, a regular density pointcloud with RGB colors and semantic labels, and a dense pointcloud with only x,y,z coordinates (with potential applications e.g. in robotics for ground traversibility mapping).

The paper describes the acquisition and processing of the dataset, then illustrates several experiments on a semantic segmentation task with a prominent point-based deep learning benchmark model (PointNet++~\cite{qi2017pointnet++}). These experiments highlight: 1) the importance of the amount of training data; 
2) the advantage of using training data from the same city on which the model is evaluated; 3) the viability of simple model ensembling approaches.

As future work, we plan to compare additional recent deep-learning models for the semantic segmentation task on this dataset. Moreover, we plan to study the effects of semi-supervised and self-supervised learning methods on unstructured pointclouds.

As we make this dataset available to the research community, we hope that it will be useful for further analysis of model generalization, domain-gap studies with respect to LiDAR datasets, and various robotics applications such as traversibility, and ultimately advance the state of the art in the field.

\section{Acknowledgments}
This work is supported partially by Nomoko AG and partially by the Swiss Confederation through Innosuisse research project 31889.1 IP-ICT and through the NCCR Robotics. The authors would like to thank the project collaborators in the SUPSI-ISIN institute 
and the colleagues in Nomoko AG for their contributions, specifically to Juan Vinuales and Mario Sanchez Gallardo for the drone flight operations; to Alexandre Ferreira do Carmo, Hugo Filipe Queiros da Cunha, Vincent Schmid, and Simon Scherer for supporting the photogrammetry steps and for manual segmentation and labeling of the pointclouds; to Sonia Batllori and Matthias Grass for supporting the pointcloud processing.


\printbibliography


\end{document}